\documentclass{article} 
\usepackage{iclr2021_conference,times}

\usepackage{amsmath,amsfonts,bm}

\def\1{\bm{1}}

\DeclareMathAlphabet{\mathsfit}{\encodingdefault}{\sfdefault}{m}{sl}
\SetMathAlphabet{\mathsfit}{bold}{\encodingdefault}{\sfdefault}{bx}{n}

\DeclareMathOperator*{\argmin}{arg\,min}

\usepackage{hyperref}
\usepackage{url}

\usepackage{booktabs}       
\usepackage{amsfonts}       
\usepackage{nicefrac}       
\usepackage{microtype}      
\usepackage{graphicx}
\usepackage{amsmath,amsfonts,amssymb,multirow,array} 
\usepackage{xcolor}
\usepackage{hyperref}
\hypersetup{colorlinks}
\usepackage{fec}
\usepackage{dpr}
\usepackage{mathtools}
\usepackage{algorithm}
\usepackage{enumerate}
\usepackage{subcaption}
\usepackage[noend]{algpseudocode}
\usepackage{makecell}
\usepackage{comment}
\usepackage{caption}
\usepackage{enumitem}
\usepackage{amsthm}
\usepackage{arydshln}
\usepackage{bm}
\usepackage{siunitx,tabularx,ragged2e,booktabs}
\usepackage{wrapfig}
\usepackage{subcaption}

\algnewcommand{\algorithmicforeach}{\textbf{for each}}
\algdef{SE}[FOR]{ForEach}{EndForEach}[1]
  {\algorithmicforeach\ #1\ \algorithmicdo}
  {\algorithmicend\ \algorithmicforeach}

\newcommand{\obproxsg}{OBProx-SG{}}
\newcommand{\hspg}{HSPG{}}
\newcommand{\algacro}{RSP{}}

\newcommand{\cifar}{\text{CIFAR10}}

\newcommand{\ilsvrc}{\text{ILSVRC2012}}
\newcommand{\imagenet}{\text{ImageNet}}
\newcommand{\resnet}{\text{ResNet50}}
\newcommand{\vgg}{\text{VGG16}}

\newcommand{\ie}{\textit{i.e.}}
\newcommand{\eg}{\textit{e.g.}}

\newcommand{\flops}{FLOPs}

\def \M {\mathcal{M}}

\def \I {\mathcal{I}}
\def \S {\mathcal{S}}

\def \O {\mathcal{O}}
\def \K {\mathcal{K}}
\def \D {\mathcal{D}}

\title{Neural Network Compression via \\Sparse Optimization}

\author{
	\textbf{Tianyi Chen}\footnotemark[1]~~\footnotemark[4]\ \ , \textbf{Bo Ji}\footnotemark[2]~~\footnotemark[4]\ \ , \textbf{Yixin Shi}\footnotemark[1]\ \ , \textbf{Tianyu Ding}\footnotemark[3]\ \ , \textbf{Biyi Fang}\footnotemark[1]\ \ , \textbf{Sheng Yi}\footnotemark[1]\ \ , \textbf{Xiao Tu}\footnotemark[1]\ \ 
}

\iclrfinalcopy 
\begin{document}

\renewcommand{\thefootnote}{\fnsymbol{footnote}}
\footnotetext[1]{Microsoft, Redmond, WA, USA. E-mail: \href{mailto:tiachen@microsoft.com}{\{tiachen,\ yixshi,\ bif,\ shengyi,\ xiaotu\}@microsoft.com}.}
\footnotetext[2]{National University of Singapore, Singapore. Email: \href{mailto:jibo@comp.nus.edu.sg}{jibo@comp.nus.edu.sg}.}
\footnotetext[3]{Johns Hopkins University, Baltimore, MD, USA. Email: \href{mailto:tding1@jhu.edu}{tding1@jhu.edu}.}
\footnotetext[4]{These authors were equally contributed.}
\renewcommand{\thefootnote}{\arabic{footnote}}

\maketitle

\begin{abstract}
The compression of deep neural networks (DNNs) to reduce inference cost  becomes increasingly important to meet realistic deployment requirements of various applications. 
There have been a significant amount of work regarding network compression, while most of them are heuristic rule-based or typically not friendly to be incorporated into varying scenarios. On the other hand, sparse optimization yielding sparse solutions naturally fits the compression requirement,  but due to the limited study of sparse optimization in stochastic learning, its extension and application onto model compression is rarely well explored. In this work, we propose a model compression framework based on the recent progress on sparse stochastic optimization. Compared to existing model compression techniques, our method is effective and requires fewer extra engineering efforts to incorporate with varying applications, and has been numerically demonstrated on benchmark compression tasks. Particularly, compared to the baseline heavy models, we achieve up to $\bm{7.2}\times$ and $\bm{2.9}\times$ FLOPs reduction with competitive $\bm{91.2\%}$ and $\bm{75.1}\%$ top-1 accuracy on~\vgg{} for~\cifar{} and \resnet{} for \imagenet{}, respectively.
\end{abstract}

\section{Introduction}

In the past decade, building deeper and larger neural networks is a primary trend for solving and pushing the accuracy boundary in varying artificial intelligent scenarios~\citep{long2015fully,vaswani2017attention}.
However, these high capacity networks typically possess significantly heavier inference costs in both time and space complexity especially when used with embedded sensors or mobile devices where computational  resources are usually limited. In addition, for 
cloud artificial intelligent service, the 
servers operating on a limited latency budget~\citep{shi2020object} benefit remarkably from the deployed models to be computationally efficient to maintain the reliable service level agreement during heavy cloud traffic. For these applications, in addition to satisfactory generalization accuracy, computational efficiency and small network sizes are crucial  factors to business success.

\paragraph{Literature Review:} There have been numerous efforts devoted to network compression~\citep{bucilu2006model} to achieve the speedup and efficient model inference, where the studies are largely evolved into \textit{(i)} weight pruning, \textit{(ii)} quantization, and \textit{(iii)} knowledge distillation. The weight pruning mainly focuses on given a heavy model with high generalization accuracy, filtering out redundant kernels to achieve slimmer architectures ~\citep{gale2019state}, including Bayesian pruning~\citep{zhou2019accelerate,neklyudov2017structured,louizos2017bayesian}, ranking  kernels importance~\citep{luo2017thinet,hu2016network,he2018soft,li2019exploiting} and compression by reinforcement learning~\citep{he2018amc}. The quantization is the process of constraining network parameters onto lower precision~\citep{wu2016quantized}. With network parameters quantized, the majority of operations, \eg, convolution and matrix multiplication, can be efficiently estimated via the
approximate inner product computation~\citep{jacob2018quantization,zhou2016dorefa}. The knowledge distillation constructs a teacher-student training pipeline to transfer the knowledge from heavy and accurate teacher network to a prescribed given smaller student network~\citep{hinton2015distilling,polino2018model}. 

We have the perspective that these existing methods tackle the network compression from different aspects and work as complementary to each other,
wherein we study the weight pruning method in this work. We note that the existing weight pruning methods are perhaps either simple and heuristic~\citep{li1608pruning,luo2017thinet}, hence easily hurt generalization accuracy; or sophisticated, so that numerous engineering efforts are required to reproduce and employ on varying applications~\citep{he2018amc}, which may be prohibitive to applications with complicated training procedures. Therefore, establishing a simple and effective compression pipeline becomes indispensable especially along with the irresistible popularization of various artificial intelligent applications. 

On the other hand, to remove redundancy of heavy models in convex learning, sparse optimization, which augments the problem with a sparsity-inducing regularization term to yield sparse solutions (including many zeros)~\citep{beck2009fast,chen2018fast}, serves as one of the most effective methods in classical feature engineering~\citep{tibshirani1996regression,roth2008group}, where the redundancy is trimmed based on the distribution of zero entries on the solution. However, its analogous applications in non-convex deep learning to filter redundancy in networks is far away from well explored. In fact, although there exist quite a few works formulating sparse optimization to compress networks, they either do not actually utilize the sparsity ratio of zero weight parameters, but rank their magnitudes to filter out weight parameters of which norm are sufficiently small~\citep{li2016pruning}; or perform not comparable to other approaches~\citep{wen2016learning,lin2019toward,gale2019state}.  As investigated in~\cite{xiao2010dual,xiao2014proximal,chen2020orthant,chen2020half}, the main reason regarding the limited performance of sparse optimization on network compression is that effective sparsity exploration in stochastic learning is hard to be achieved due to the randomness and limited sparsity identification mechanism, \ie, the solutions computed by most of stochastic sparse optimization algorithms are typically fully dense. Hence, the previous compression techniques by sparse optimization may not utilize the accurate sparsity signal to achieve effective model compression.

Moreover, the sparsity identification in stochastic non-convex learning are improved to some extent by the recent progress made by~orthant-face and half-space projection for~$\ell_1$-regularization and group sparsity regularization problems respectively~\citep{chen2020orthant,chen2020half}. Therefore, it is natural to ask if the recent progress on stochastic sparse optimization is applicable and beneficial to the network compression scenario. In this work, we attempt to re-investigate the utility of sparse optimization on the weight pruning task to search approximately optimal model architectures.


\begin{figure*}[t]
\centering
\includegraphics[scale=0.69]{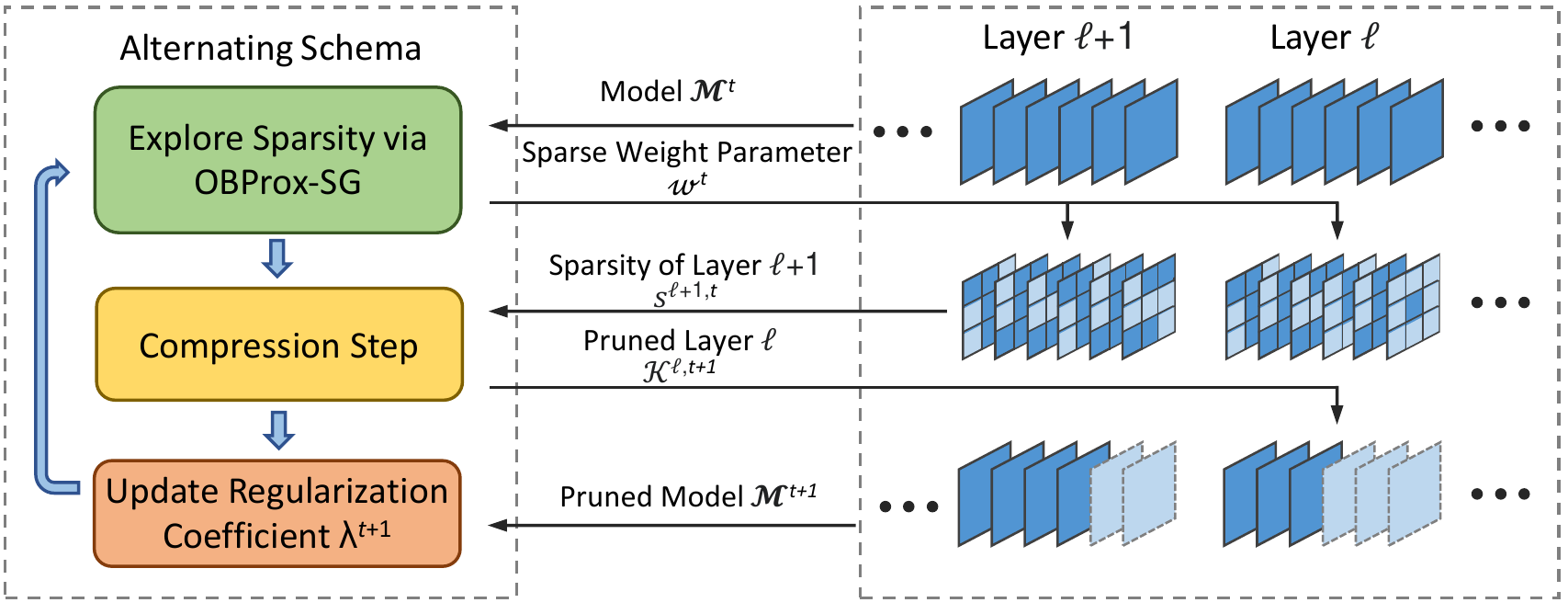}
\caption{Overview of alternating schema.}
\label{fig.overview}
\end{figure*}

\paragraph{Our Contributions:}

We propose a novel network compression framework based on the recent sparse optimization algorithm~\obproxsg{} for solving $\ell_1$-regularization, so-called Recursive Sparse Pruning (\algacro) method. Compared to other weight pruning methods, \algacro{} is easily plugged into various applications with competitive performance. We now summarize our contributions as follows.
\begin{itemize}[leftmargin=*]
    \item \emph{Algorithmic Design:} We formulate the model compression problem into an optimization problem,
    and design an alternating recursive schema as shown in Figure~\ref{fig.overview} to search an approximately optimal solution wherein the optimal model architecture requires much fewer time and space cost for inference and achieves competitive accuracy to the original full model.
    Compared to other pruning methods,~\algacro{} is more friendly to incorporate with various applications with much fewer efforts, as the fundamental herein step, \ie, redundancy exploration, is tackled by a well-packaged optimization algorithm~(\obproxsg{}) applicable for broad training pipelines. 
    \item \emph{Numerical Experiments:} We numerically demonstrate the effectiveness of our compression approach on two benchmark model compression problems, \ie,~\vgg{} on~\cifar{} and~\resnet{} on~\imagenet{} (\ilsvrc{}) with up to $7.2\times$ and $2.9\times$ FLOPs reduction and only slight accuracy regression. Compared to the results presented in~\cite{zhou2019accelerate}, our approach outperforms all other pruning methods on both~\flops{} reduction and accuracy maintenance significantly. 
\end{itemize}

\section{Notations and Preliminaries}

Consider a CNN model consisting of $L$ convolutional layers, which are interlaced with various activation functions, \eg, rectifier linear units, and pooling operators, \eg, max pooling. For the convolution operation on $l$-th convolutional layer, an input tensor $\I^{l}$ of size $ C^{l}\times H^{l}_{\I} \times W^{l}_{\I}$, is transformed into an output tensor $\O^{l}$ of size $C^{l+1}\times H^{l}_{\O} \times W^{l}_{\O}$ by the following linear mapping:
\begin{equation}
\O^{l}_{c',h',w'}=\sum_{c=1}^{C_l}\sum_{i=1}^{H_{\K}^l}\sum_{j=1}^{W_{\K}^l}\mathcal{K}^{l}_{c',c,i,j}\I_{c, h_i, w_j}^l,
\end{equation}
where the convolutional filter $\K^{l}$ at the $l$th layer is a tensor of size $K^{l}\times C^{l}\times H_{\K}^l\times W_{\K}^l$ with $K^{l}=C^{l+1}$. Particularly, we denote $\K^{l}_{c'}\in\mathbb{R}^{C^{l}\times H_{\K}^l\times W_{\K}^l}$ as the $c'$-th kernel for the $l$-th convolutional layers. 
Similarly to~\cite{lin2019toward}, we flatten tensor $\K^{l}$ into a filter matrix $\K^{l}$ of size $K^{l} \times {C^{l}H_{\K}^l W_{\K}^l}$. Hence, the $c'$-th row of $\K^{l}$ exactly store the parameters for the $c'$-th kernel. 

Additionally, we consider several norms of the filter matrix $\K^{l}$ that are widely used to formulate sparsity inducing regularization problem. Particularly, one is the $\ell_1$-norm of $\K^{l}$ as
\begin{equation}
\norm{\K^{l}}_1=\sum_{c'}\norm{\K^{l}_{c'}}_1,
\end{equation} 
which induces a sparse solution while the zero elements may be distributed randomly. To encode more sophisticated group sparsity structure, mixed $\ell_1/\ell_2$ norm is introduced as 
\begin{equation}
\norm{\K^{l}}_{1,2}=\sum_{c'}\norm{\K^{l}_{c'}}_2
\end{equation}
which induces the entire rows of $\K^{l}$ to be zero during solving sparse optimization problem. In this work, we will focus on leveraging the sparse $\ell_1$-regularization optimization to compress heavy networks and pay attention to mixed $\ell_1/\ell_2$-regularization to encode group sparsity in the future work.

\section{Network Compression Problem Formulation}

Essentially, the network compression is to search a slimmer model architecture given a prescribed network topology along with minimizing the scarification on generalization accuracy, which can be formulated as the following constrained optimization problem,
\begin{equation}\label{prob.main.raw}
\begin{split}
\minimize{\bm{w}, \M}&\ f(\bm{w}~|~\M,\D)\\
s.t.&\ |\M|\leq T,
\end{split}
\end{equation}
where $\M$ denotes the model architecture associated with its weight parameter $\bm{w}$, $f$ is the loss function to measure the deviation from the predicted output given evaluation dataset $\D$ on $(\bm{w}, \M)$ to the ground truth output, and the constraint $|\M|\leq T\in\mathbb{Z}^+$ restricts the size of model $\M$. It is well-recognized that~\eqref{prob.main.raw} is NP-hard and intractable in general. But one can solve it approximately by transforming into non-constraint optimization problem as 
\begin{equation}\label{prob.main.intermediate}
\minimize{\bm{w}, \M}\ f(\bm{w}~|~\M,\D) + \lambda \cdot \Omega(\mathcal{M}),
\end{equation}
where the constraint in~\eqref{prob.main.raw} is augmented into the original objective as an extra plenaty term $\Omega(\mathcal{M})$ associated with a Lagrangian coefficient $\lambda$. The $\Omega$ serves as a regularization regarding the size of model $\M$, which can be further relaxed to the $\ell_1$-norm of weight parameters $\bm{w}$ on $\M$,
\begin{equation}\label{prob.main}
\minimize{\bm{w}, \M}\  f(\bm{w}~|~\M,\D) + \lambda \cdot \norm{\bm{w}}_1.
\end{equation}
Compared to~\eqref{prob.main.raw} and~\eqref{prob.main.intermediate}, problem~\eqref{prob.main} is tractable by our following proposed alternating algorithm, where we iteratively update $\bm{w}$ given fixed $\M$, then construct $\M$ given computed $\bm{w}$ till convergence. The solution $(\bm{w}^*, \M^*)$ of problem~\eqref{prob.main} are expected to possess smaller model size and similar evaluation accuracy compared to the initial heavy model $\M^0$ to accelerate the inference efficiency.

\section{The Compression Method}\label{sec.algorithm}

In this section, we provide a comprehensive introduction of our compression technique via sparse optimization as illustrated by Figure~\ref{fig.overview}. Our \algacro{} prunes redundant filters from a high capacity model for computational efficiency while avoiding sacrificing the generalization accuracy. In general, to solve the formulated compression problem~\eqref{prob.main}, our~\algacro{} as stated in Algorithm~\ref{alg:main} is a framework consisting of two stages, \textit{(i)} an iteratively alternating compression schema to search an approximately optimal model architecture, \ie, iteratively repeat updating $\bm{w}^t$ given $\M^t$ then updating $\M^{t+1}$ given $\bm{w}^t$  till convergence as line~\ref{line.alternating.start} to~\ref{line.alternating.end} in Algorithm~\ref{alg:main}; and \textit{(ii)} fine-tune model parameters given searched optimal model architecture as line~\ref{line.global_fine_tuning} in Algorithm~\ref{alg:main}.  In the reminder of this section, we first present how the alternating compression schema works, then move on to explain the fine-tuning of network weight parameters, and conclude this section by revealing the Bayesian interpretation of our approach.  

\begin{algorithm}[H]
	\caption{The Network Compression via Sparse Optimization Framework}
	\label{alg:main}
	\begin{algorithmic}[1]
		\State \textbf{Input:} Initial high capacity CNN architectures $\mathcal{M}^0$ and dataset $\mathcal{D}$.
		\While{Network compression does not converge.}\label{line.alternating.start}
		\State \textbf{Update $\bm{w}^{t}$ given $\M^t$}: Train weight parameter $\bm{w}^{t}$ on dataset $\mathcal{D}$ as 
			\begin{align*}
				\bm{w}^{t} \gets \text{SparseOptimization}(\M^t, \mathcal{D},\lambda^{t}), \tag{Explore redundancy} 
			\end{align*}\label{line.obproxsg}
		\State \textbf{Update $\M^{t+1}$ given $\bm{w}^t$}: Construct model $\mathcal{M}^{t + 1}$ by Algorithm~\ref{alg:compress.general}, 
		\begin{equation}
		\mathcal{M}^{t + 1}\gets \text{Compression}(\M^t, \bm{w}^t) \tag{Compress $\M^t$}
		\end{equation}
		\State \textbf{Update $\lambda^{t+1}$}: $\lambda^{t+1}$ computes proportionally to the model size ratio between $\M^{t+1}$ and $\M^t$,
		\begin{equation}
		\lambda^{t+1} \gets |\mathcal{M}^{t+1}|~/~|\mathcal{M}^{t}|\cdot \lambda^{t}.   
		\end{equation} \label{line.alternating.end}
		\EndWhile
		\State \textbf{Global Fine Tuning:} Retrain to attain weights $\bm{w}^*$ given the optimal model $\mathcal{M}^*$. \label{line.global_fine_tuning}
		\State \textbf{Output:} Pruned CNN architecture $\mathcal{M}^*$ with tuned weight parameters $\bm{w}^*$.
	\end{algorithmic}
\end{algorithm}


\paragraph{Explore Redundancy, update $\bm{w}^t$ given $\M^t$.} Given the current possibly heavy model architecture $\M^{t}$, we aim at evaluating its redundancy of each layer in order to construct smaller structures with the same prediction power. To achieve it, we first formulate and train $\M^{t}$ under a sparsity-inducing regularization to attain a highly sparse solution $\bm{w}^t$, \ie,
\begin{equation}\label{prob.l1reg}
\bm{w}^{t} =\argmin_{\bm{w}}F\left( \bm{w} ~|~ \mathcal{M}^{t},\D \right) := f(\bm{w} ~|~ \mathcal{M}^{t},\D) + \lambda^{t} \cdot \norm{\bm{w}}_1,
\end{equation}
where the $\norm{\bm{w}}_1$ penalizes on the density and magnitude of the network weight parameters; theoretically, under proper $\lambda^t$, the solution of problem~\eqref{prob.l1reg} $\bm{w}^{t}$ tends to have both high sparsity and evaluation accuracy. The problem as form of~\eqref{prob.l1reg} is well-known as $\ell_1$-regularization problem, which is widely used for feature selection in convex applications to filter out redundant features by the sparsity of solutions.  Consequently, it is natural to consider to apply such regularization to determine redundancy of deep neural networks. Unfortunately, although problem as form of~\eqref{prob.l1reg} has been well studied in convex deterministic learning~\citep{beck2009fast,keskar2015second,chen2017reduced}, its study in stochastic non-convex applications, \eg, deep learning, is quite limited, \ie, the solutions computed by existing stochastic optimization are typically dense. Such limitation on sparsity exploration results in that the $\ell_1$-regularization is typically not used in a standard way to identify redundancy of network compression, where it is leveraged to rank the importance of filters only in the aspect of the magnitude but not on the sparsity in the recent literatures~\citep{li2016pruning}. 

We note that a recent orthant-face prediction method (\obproxsg)~\citep{chen2020orthant} made a breakthrough on the study of $\ell_1$-regularization especially on the sparsity exploration compared to classical proximal methods in stochastic non-convex learning~\citep{xiao2010dual,xiao2014proximal}. Particularly, the solutions computed by~\obproxsg{} typically have multiple-times higher sparsity and competitive objective convergence than the solutions computed by the competitors. Hence, we employ the~\obproxsg{} onto line~\ref{line.obproxsg} in Algorithm~\ref{alg:main} to seek highly sparse weight parameters $\bm{w}^t$ on model architecture $\M^t$ without accuracy regression. 

\paragraph{Compress model, update $\bm{M}^{t+1}$ given $\bm{w}^t$ and $\M^t$.}

Given the highly sparse weight parameters $\bm{w}^t$ of model architecture $\M^t$ yielded by~\obproxsg{}, as the high-level procedure presented in Algorithm~\ref{alg:compress.general}, we now perform a specified compression algorithm, \eg, filter pruning, to construct a smaller model architecture $\M^{t+1}$, wherein the sparsity ratio of each building block is leveraged to represent the redundancy of corresponding layers.  Particularly, taking the fully convolution layers (no residual or skip connection) as a concrete example, as stated in Algorithm~\ref{alg:compress.fullcnn}, the sparsity ratio of each $\ell$-th convolutional layer is computed as line~\ref{line.compute_sparsity.start} to~\ref{line.compute_sparsity.end} by counting the percentage of zero elements. Then following the backward path of $\M^{t+1}$, the $\ell$-th convolutional layer shrinks its number of filters proportionally to computed sparsity ratio $s^{\ell+1, t}$ as line~\ref{line.shrink_filters.start} to~\ref{line.shrink_filters.end}.
 Finally the architecture of $\M^{t+1}$ is finalized after a calibration on the potential structure inconsistency during pruning to guarantee the ultimate network generating outputs in the desired shape.  

\paragraph{Update the regularization coefficient $\lambda^{t+1}$ given $\lambda^t, \M^t$ and $\M^{t+1}$.} The regularization coefficient $\lambda$ in problem~\eqref{prob.main} balances between the sparsity level on the exact optimal solution and the generalization accuracy. Larger $\lambda$ typically results in higher sparsity on the computed solution $\bm{w}$ by~\obproxsg{} but sacrifices more on the bias of model estimation. Hence, to achieve both low objective  value and high sparse solution, $\lambda$ requires careful selection, of which setting is usually related to the number of variables involved per iteration during optimization. Thus, we scale $\lambda^t$ up according to the model size ratio between $\M^{t+1}$ and $\M^t$ to yield $\lambda^{t+1}$ as line~\ref{line.alternating.end} in Algorithm~\ref{alg:main}.

\begin{algorithm}[h]
	\caption{Compress Step for General CNNs}
	\label{alg:compress.general}
	\begin{algorithmic}[1]
		\State \textbf{Input:} CNN model $\mathcal{M}^t$ of $L^{t}$ building blocks, sparse weights $\bm{w}^t$.
		\For{$\ell=1,\cdots, L^{t}$}\label{line.compute_sparsity.start}
		    \State Compute the empirical sparsity ratio $s^{\ell,t} $ given $\bm{w}^t$ as 
		    \begin{equation}
		      s^{\ell,t}\gets \frac{\text{\# of zeros in}\ \K^{l,t} }{\text{\# of elements in}\ \K^{l,t}}
		    \end{equation}\label{line.compute_sparsity.end}
		\EndFor
		\State Remove redundancy in each building block by sparsity ratio $\{s^{\ell,t}\}_{\ell=1}^{\ell=L^t}$ to form $\M^{t+1}$. 
 		\State \textbf{Output:} The CNN model $\mathcal{M}^{t + 1}$.
	\end{algorithmic}
\end{algorithm}

\begin{algorithm}[h]
	\caption{Compress Step for Fully Convolutional Layers}
	\label{alg:compress.fullcnn}
	\begin{algorithmic}[1]
		\State \textbf{Input:} Parameter $\bm{x}^t$, CNN model $\mathcal{M}^t$ of $L^{t}$ convolutional layers and weight parameters $\bm{w}^t$.
		\For{$\ell=1,\cdots, L^{t}$}\label{line.compute_sparsity.start}
		    \State Compute the empirical sparsity ratio $s^{\ell,t} $ given $\bm{w}^t$ as 
		    \begin{equation}
		      s^{\ell,t}\gets \frac{\text{\# of zeros in}\ \K^{l,t} }{\text{\# of elements in}\ \K^{l,t}}
		    \end{equation}\label{line.compute_sparsity.end}
		\EndFor
		\State Initialize $\mathcal{M}^{t + 1}$  with $L^{t+1}\gets L^t$ convolution layers of the same architectures as $\M^t$:
		\For{$\ell=L^{t+1}-1,\cdots, 1$}\label{line.shrink_filters.start}
		    \State In $\ell$-th convolution layer, shrink the number of filters from $K^{\ell, t}$ to
				\begin{equation}
				K^{\ell, t+1}=\left\lceil K^{\ell, t} \cdot (1-{s}^{\ell + 1}) \right\rceil
				\end{equation}\label{line.shrink_filters.end}
		\EndFor
		\State Calibrate the architecture of $\M^{t+1}$. \label{line.calibrate_structure} 
 		\State \textbf{Output:} The CNN model $\mathcal{M}^{t + 1}$.
	\end{algorithmic}
\end{algorithm}

\paragraph{Convergence Criteria} The stopping criteria of the alternating schema is designed delicately to ensure obtaining the approximate minimal model architecture without accuracy regression.  Our design monitors the several evaluation metric, \eg, sparsity, model size and validation accuracy, during the whole procedure. In particular, if the evolution of sparsity of $\bm{w}$ and model size $\M$ becomes flatten or the validation accuracy regresses, the alternating schema is considered as convergent and return the best checkpoint as the optimal model $\M^*$ as line~\ref{line.alternating.start} in Algorithm~\ref{alg:main}. 
 
\paragraph{Global Fine-Tuning} After searching the optimal model $\M^*$, of which size tends to be much smaller than the original $\M^0$. In the final step, we retrain the network to strengthen the remaining neurons after compression to enhance the generalization performance of the trimmed network as line~\ref{line.global_fine_tuning} in Algorithm~\ref{alg:main}.

\paragraph{Bayesian Interpretation} It is well recognized that $\ell_1$-regularization in problem~\eqref{prob.l1reg} is equivalent to adding a Laplace prior onto the training problem to enhance the occurrence of zero entries on the solutions. From this point of view, our method can be categorized into weight pruning by Bayesian compression, where other methods constructs other priors into the original objectives~\citep{zhou2019accelerate,louizos2017bayesian,neklyudov2017structured}.

\section{Numerical Experiments}\label{sec:exp}

In this section, we validate the effectiveness of the proposed method~\algacro{} on two benchmark datasets~\cifar{}~\citep{Krizhevsky09} and~ImagetNet~(\ilsvrc{})~\citep{deng2009imagenet}. The CNN architectures to prune include~\vgg~\citep{simonyan2014very} and~\resnet~\citep{he2016deep}. We mainly report floating operations (FLOPs) to indicate acceleration effect. Compression rate (CR) is also revealed as another criterion for pruning. 

\paragraph{Experimental Settings} 
For~\obproxsg{} called in~Algorithm~\ref{alg:main}, the initial regularization coefficient $\lambda^0$ is fine-tuned from all powers of 10 between $10^{-2}$ to $10^{-5}$ as the one can achieve the same generalization accuracy as the existing benchmark accuracy. The mini-batch sizes for~\cifar{} and~\ilsvrc{} are selected as 64 and 512 respectively The learning rates follow the setting as~\cite{he2016deep} and decay periodically by factor $0.1$. In practice, due to the expansiveness of training cost, we empirically set the number of alternating update as one.

\paragraph{\vgg{} on~\cifar{}} We first compress~\vgg{} on~\cifar{}, which contains $50,000$ images of size $32\times 32$ in training set and $10,000$ in test set. The model performance on~\cifar{} is qualified by the accuracy of classifying ten-class images. Considering that~\vgg{} is originally proposed for large scale dataset, the redundancy is naturally obvious. In realistic implementation, we add a lower bound of compression ratio,~\ie, $\epsilon>0$, into~line~\ref{line.shrink_filters.end} of Algorithm~\ref{alg:compress.fullcnn}, such that   
\begin{equation}
\begin{split}
K^{\ell, t+1}=
\left\lceil K^{\ell, t} \cdot \max\{1-{s}^{\ell + 1}, \epsilon\} \right\rceil
\end{split}
\end{equation}
to guarantee each layer maintain a fair number of kernels left after compression and select $\epsilon$ as 0.1. 
We report the pruning results in Table~\ref{table:vgg_cifar}, including existing state-of-the-art compression techniques reported in~\cite{zhou2019accelerate}.  
Particularly, Table~\ref{table:vgg_cifar} shows that~\algacro{} is no doubt the best compression method on the FLOPs reduction, by achieving $7.2\times$ speedups whereas others stuck on about $3\times$ speedups. Remark here that although \algacro{} reaches slight lower compression ratio than RBP because of the universal setting of $\epsilon$  for both convolution and linear layers, the~\flops{} reduction of~\algacro{} is roughly 2 times higher due to the more significant computational acceleration on the early layers. 
 If we switch the gear to compare the baseline validation accuracy,~\algacro{} only slightly regressed about $0.4\%$, which outperforms the majority of other compression techniques as well.

\begin{table}[t]
	\centering
	\caption{Comparison of pruning~\vgg{} on~\cifar{}. Convolutional layers are in bold.}
	\label{table:vgg_cifar}
	\resizebox{\textwidth}{!}{
		\begin{tabular}{ 
				c|c|c|c|c}
			\Xhline{3\arrayrulewidth}
		    Method & Architecture & CR & FLOPs & ACC \\
		    \hline
		    Baseline & \textbf{64-64-128-128-256-256-256-512-512-512-512-512-512}-512-512 & $1.0\times$ & $1.0\times$ & 91.6 \\
		    SBP &  \textbf{47-50-91-115-227-160-50-72-51-12-34-39-20}-20-272 & $17.0\times$ & $3.2\times$ & 91.0 \\
		    BC &  \textbf{51-62-125-128-228-129-38-13-9-6-5-6-6}-6-20 & $18.6\times$ & $2.6\times$ & 91.0 \\
		    RBC & \textbf{43-62-120-120-182-113-40-12-20-11-6-9-10}-10-22 & $25.7\times$ & $3.1\times$ & 90.5 \\
		    IBP & \textbf{45-63-122-123-246-199-82-31-20-17-14-14-31}-21-21 & $13.3\times$ & $2.3\times$ & 91.7\\
		    RBP & \textbf{50-63-123-108-104-57-23-14-9-8-6-7-11}-11-12 & $39.1\times$ & $3.5\times$ & 91.0\\
		    \textbf{\algacro{}} &\textbf{55-31-65-63-115-75-43-52-52-52-52-52-52}-52-52 & $38.9\times$ & $\bm{7.2\times}$ & 91.2 \\
			\hline
			\Xhline{3\arrayrulewidth} 
	\end{tabular}}
	
	\vspace{-0.1in}
\end{table} 

\paragraph{\resnet{} on~\ilsvrc{}:} We now prune~\resnet{} on~\ilsvrc{}~\citep{deng2009imagenet}. \ilsvrc{}, well known as ImageNet, is a large-scale image classification dataset, which contains $1,000$ classes, more than $1.2$ million images in training set and $50,000$ in validation set. \resnet{} is a very deep CNN in the residual network family. It contains 16 residual blocks~\citep{he2016deep}, where around 50 convolutional layers are stacked. For ResNet, there exist some restrictions due to its special structure. For example, the channel number of each block in the same group needs to be consistent in order to finish the sum operation. Thus it is hard to prune the last convolutional layer of each residual block directly. Hence, we consider the residual block as the basic unit rather than convolutional layer to proceed compression during Algorithm~\ref{alg:compress.general}. Particularly, the sparsity ratio of each residual block is computed and then leveraged to prune redundant filters similarly to the strategy in~\cite{luo2017thinet}. The compression results is described in Table~\ref{table:resnet_imagenet}, where the~\algacro{} definitely performs the best, with remarkable FLOPs reduction and highest Top 1 and 5 accuracy. 

\begin{table}[t]
	\centering
	\caption{Comparison of pruning~\resnet{} on~\ilsvrc{}.}
	\label{table:resnet_imagenet}
		\begin{tabular}{ c|c|c|c}
			\Xhline{3\arrayrulewidth}
		    Method & FLOPs & Top-1 Acc. & Top-5 Acc. \\
		    \hline
		    Baseline & $1.0\times$ & $76.1$ & $92.9$ \\
		    DDS & $1.7\times$ & $71.8$ & $91.9$ \\
		    CP & $1.5\times$ & $72.3$ & $90.8$ \\
		    ThiNet-50 & $2.3\times$ & $71.0$ & $90$ \\
		    RBP & $2.3\times$ & $71.1$ & $90.0$\\
		    \textbf{\algacro{}} & $\bm{2.9\times}$ & $\bm{75.1}$ & $\bm{91.9}$ \\
			\hline
			\Xhline{3\arrayrulewidth} 
	\end{tabular}
	
	\vspace{-0.1in}
\end{table} 

\section{Conclusions and Future Work}

In this report, we propose a model compression framework based on the $\ell_1$-regularized sparse optimization and the recent breakthrough on sparsity exploration achieved by~\obproxsg. Our~\algacro{} contains a recursive alternating schema to search an approximately optimal model architecture to accelerate network inference without accuracy scarification by leveraging the computed sparsity distribution during algorithmic procedure. Compared to other Bayesian weight pruning methods, the~\algacro{} is no doubt the best in terms of FLOPs reduction and accuracy maintenance on benchmark compression experiments, \ie, \resnet{} on~\imagenet{} by reducing~\flops{} by $2.9\times$ and achieving $75.1\%$ top-1 accuracy. As the future work, we remark here that how to utilize the recent group-sparsity optimization algorithm~\hspg{}~\citep{chen2020half} is an open problem to remove entire redundant hidden structures from networks directly. 

\bibliography{paper}
\bibliographystyle{iclr2021_conference}

\end{document}